\documentclass[review]{elsarticle}

\usepackage[utf8]{inputenc}
\usepackage[T1]{fontenc}
\usepackage{booktabs}
\usepackage{graphics}
\usepackage{verbatim}
\usepackage{bm}
\usepackage{CJKutf8}
\usepackage{ulem}
\usepackage{amsmath}
\usepackage{amsfonts}
\usepackage{multirow}
\usepackage[dvipsnames]{xcolor}
\usepackage[section]{placeins}

\newcommand{\etal}{\textit{et al.}}

\renewcommand{\sout}[1]{\unskip}

\journal{Journal of Biomedical Informatics}

\begin{document}

\begin{frontmatter}

\title{Identification of Pediatric Respiratory Diseases Using Fine-grained Diagnosis System}

\author[add1,add6]{Gang Yu\fnref{contribute}}
\author[add4]{Zhongzhi Yu\fnref{contribute}}
\author[add3]{Yemin Shi\fnref{contribute}}
\fntext[contribute]{Contributed equally to this work.}

\author[add2,add6]{Yingshuo Wang}
\author[add4]{Xiaoqing Liu}
\author[add1,add6]{Zheming Li}
\author[add1,add6]{Yonggen Zhao}
\author[add4]{Fenglei Sun}
\author[add5]{Yizhou Yu\corref{mycorrespondingauthor}}
\cortext[mycorrespondingauthor]{Corresponding authors at: Department of Computer Science, The University of Hong Kong, Pokfulam, Hong Kong (Y. Yu). No. 3333 Binsheng Road, Hangzhou, Zhejiang, PRChina (Q. Shu)}
\ead{yizhouy@acm.org}
\author[add6]{Qiang Shu\corref{mycorrespondingauthor}}
\ead{shuqiang@zju.edu.cn}

\address[add1]{Department of IT Center, the Children's Hospital, Zhejiang University School of Medicine, China}
\address[add2]{Department of Pulmonology, The Children’s Hospital, Zhejiang University School of Medicine, China}
\address[add3]{Department of Computer Science, School of EE\&CS, Peking University, Beijing, China}
\address[add4]{Deepwise AI Lab, Beijing, China}
\address[add5]{Department of Computer Science, The University of Hong Kong}
\address[add6]{National Clinical Research Center for Child Health, China}

\begin{abstract}
Respiratory diseases, including asthma, bronchitis, pneumonia, and upper respiratory tract infection (RTI), are among the most common diseases in clinics. 
The similarities among the symptoms of these diseases precludes prompt diagnosis upon the patients' arrival. 
In pediatrics, the patients' limited ability in expressing their situation makes precise diagnosis even harder. 
This becomes worse in primary hospitals, where the lack of medical imaging devices and the doctors' limited experience further increase the difficulty of distinguishing among similar diseases. 
In this paper, a pediatric fine-grained diagnosis-assistant system is proposed to provide prompt and precise diagnosis using solely clinical notes upon admission, which would assist clinicians without changing the diagnostic process.
The proposed system consists of two stages: a test result structuralization stage and a disease identification stage.
The first stage structuralizes test results by extracting relevant numerical values from clinical notes, and the disease identification stage provides a diagnosis based on text-form clinical notes and the structured data obtained from the first stage.
A novel deep learning algorithm was developed for the disease identification stage, where techniques including adaptive feature infusion and multi-modal attentive fusion were introduced to fuse structured and text data together.
Clinical notes from over 12000 patients with respiratory diseases were used to train a deep learning model, and clinical notes from a non-overlapping set of about 1800 patients were used to evaluate the performance of the trained model.
The average precisions (AP) for pneumonia, RTI, bronchitis and asthma are 0.878, 0.857, 0.714, and 0.825, respectively, achieving a mean AP (mAP) of 0.819.
These results demonstrate that our proposed fine-grained diagnosis-assistant system provides precise identification of the diseases.

\end{abstract}


\end{frontmatter}

% \linenumbers

\section{Introduction}
Respiratory diseases are the set of diseases that affect the organs and tissues in the human respiratory system, making gas exchange difficult~\cite{tortora2008principles}.
Complaints about respiratory disease symptoms, such as cough and mucus, are among the most common symptoms encountered in medicine~\cite{umoh2013pattern}.
The morbidity and mortality rates of respiratory diseases are among the top ones around the world~\cite{world2008global,british2006burden, hubbard2006burden,schraufnagel2013official,gibson2013respiratory,world2006country,ramanakumar2005respiratory}.
The situation is severe in developing countries, where respiratory diseases are the leading cause of death~\cite{world2007global}. 
One of the major causes of respiratory diseases is environmental exposure, which often means breathing unclean air~\cite{schraufnagel2013official, fowler2010official, slatore2010official}. 
Children's limited immunity makes the situation even worse in pediatrics. 
In 2017 alone, about 0.8 million children under five years died of pneumonia and other lower respiratory infections (LRI), accounting for 15\% of total deaths~\cite{child_death}. 

Among the respiratory diseases, pneumonia, RTI, bronchitis, and asthma have high morbidity. 
However, the similarity among the symptoms of these diseases make prompt diagnosis in clinics upon admission challenging, since typically only limited information from inquiry and simple tests is available.
All four diseases share the symptom of coughing.
In most cases, asthma patients have symptoms of wheezing, although pneumonia and bronchitis patients may also experience wheezing.
In most cases, chest radiography is necessary to achieve accurate diagnosis.
However, according to a study by the World Health Organization (WHO) and the Pan America Health Organization (PAHO), two-thirds of the world's population have no access to medical imaging including X-rays~\cite{paho2012world}.
The limited experience of doctors at primary hospitals also worsens the situation. 
Thus, misdiagnosis of these diseases is common at primary hospitals.
Inappropriate treatment as a result of misdiagnosis leads to prolonged recovery and potential exacerbation~\cite{nantanda2013asthma}.
It is crucial to develop an accurate and prompt system to assist the diagnosis of the aforementioned respiratory diseases in less developed regions. 

The situation is even worse in pediatrics. 
Children have limited ability to clearly express themselves, especially when suffering from pain, which increases the challenge of accurate clinical diagnosis upon arrival.
The commonly used approach of hypothetico-deductive reasoning is not very effective as the answer provided by the patient might not correctly reveal his/her situation. 
Doctors often need to diagnose patients through their own observations and the chief complaints provided by the patients' parents. 
Hence, to function properly in pediatrics, the system should be able to perform on doctors' observations and limited assistance from parents. 
To overcome children's limited self-expression skills, in our proposed diagnosis system, we only use chief complaints and the outcomes of clinical tests and physical examinations. Such information can be obtained through observation and examinations and does not require hypothetico-deductive reasoning.

\begin{figure}
    \centering
    \includegraphics[width=\textwidth]{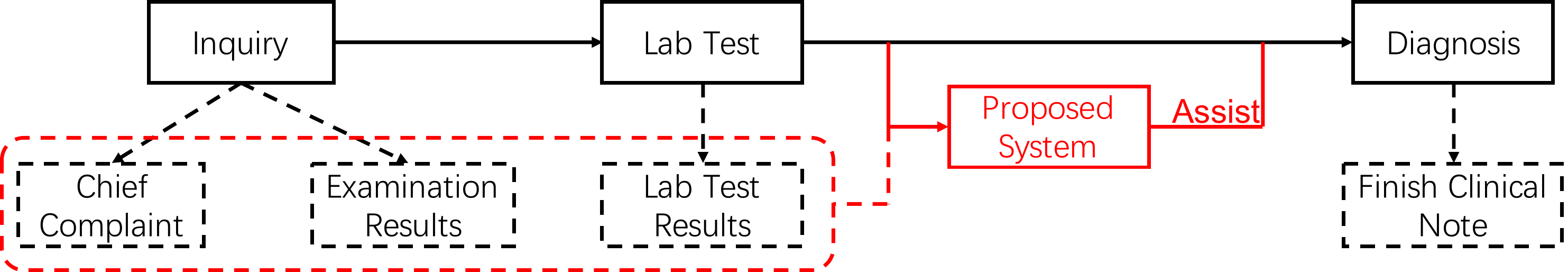}
    \caption{The original diagnosis procedure and the procedure using the proposed system (better view in color version). Black solid line and boxes are original procedure, black dashed lines and boxes are generated clinical notes at each step. Red solid lines are the additional step of using the proposed system while red dashed box and line are the existing clinical notes used by the proposed system.}
    \label{fig:diagnosis flow}
\end{figure}

In this study, we first explore the possibility of effectively analyzing unstructured clinical notes for identifying respiratory diseases and test result structuralization without exploiting any manual annotations.  
Then, a pediatric fine-grained diagnosis-assistant system is proposed to identify four respiratory diseases, including pneumonia, RIT, bronchitis and asthma.
To address the lack of high-quality structured data, our system takes the original unstructured clinical notes as the input, avoiding any format conversion or manual annotations. 
These clinical notes include chief complaints, physical examinations and test results. 
The diagnosis procedure comparison between using the proposed system or not is shown in Fig.~\ref{fig:diagnosis flow}. It can be clearly seen that adopting the proposed system provides reliable assistance without requiring any changes to the original diagnosis procedure. Specifically, given a clinical note, the proposed fine-grained diagnosis-assistant system can make a multi-label prediction, output the possible respiratory diseases to help the clinician with the diagnosis.
As no additional data, e.g. administrative data, is used except for clinical notes, the proposed method can also be used for quality control and auditing purposes so as to provide a warning for any serious diagnosis errors.

The proposed system consists of two stages: a test result structuralization stage and a disease identification stage.
As shown in previous work, test results are important evidence, and numerical values in such test results can significantly help with diagnosis.
To make full use of test results, the first structuralization stage fills numerical fields in the structured part of the data by applying a set of regular expressions as templates for extracting test results from clinical notes in an unsupervised manner, giving rise to efficient and light-weight test results structuralization.
It is worth mentioning that such extracted structured data still has high sparsity as the majority of patients take a limited number of tests only. To reduce the level of sparsity, the extracted structured data is further pruned before being passed to the disease identification stage.

The disease identification stage takes a combination of the text-form clinical note (text data) and the structuralized test results as the input.
To address similarities among symptoms of different respiratory diseases, we propose an infusion block and a multi-modal attentive fusion module to fuse structured and text data and extract fine-grained features in the disease identification stage.
It is easy to understand that symptoms in chief complaints are related to the test results, and their connections can help reveal the target disease.
It is also known that different depths in a deep neural network have observations of the input data at different scales. Therefore, we first employ an adaptive feature infusion block to infuse structured data into text data at multiple depths of our network so as to introduce multi-scale connections.
Then, the multi-modal attentive fusion module is introduced to compute dual attention between structured and text data, achieving feature fusion.

To evaluate the proposed system, 14,697 clinical notes from patients with at least one of the four aforementioned respiratory diseases are collected. 11,100 samples are used to train the system, and 1,821 samples are used as a validation set. Finally, the rest of the 1,776 samples form a test set for evaluating the performance of the system. The average precisions (AP) on the test set for pneumonia, RTI, bronchitis and asthma are 0.878, 0.857, 0.714, and 0.825, respectively, resulting in a mean AP (mAP) of 0.819. The results on the test set demonstrate that our proposed fine-grained diagnosis-assistant system achieves precise disease identification.

\section{Background and Related Works}

Given the characteristics of pediatrics, the pediatric-specific disease prediction systems using neural language processing (NLP) or machine learning have not yet been extensively exploited. 
Aczon \etal~\cite{aczon2017dynamic} and Ehwerhemuepha \etal~\cite{ehwerhemuepha2018novel} tried to predict the course of disease from pre-processed electronic health records (EHR) data using the logistic regression and recurrent neural network (RNN), respectively. 
Neuman \etal~\cite{neuman2011prediction} focused on the diagnosis in the pediatric, but they needed the radiography information to make the prediction, making it infeasible to resource-constrainted hospitals. 
Furthermore, in 2020, Roquette \etal~\cite{roquette2020prediction} proposed to directly analyze unstructured text data using a 2-layer deep neural network to predict admission in pediatric, which represents the state-of-the-art performance in the applying machine learning and neural language processing in pediatric.
There were also some works trying to introduce multi-modal information into the diagnosis system. 
A novel Text-Image Embedding network (TieNet) was proposed by Wang \etal~\cite{wang2018tienet} to classify the chest X-rays and generate reports via extracting distinctive information from both image and text representations.
To diagnose based on both text and image information, Shin \etal~\cite{shin2015interleaved} also introduced an interleaved text/image deep learning system to extract and mine the semantic interactions of radiology images and reports. Liang \etal~\cite{liang2019evaluation} proposed machine learning classifiers (MLCs) that could query electronic health records (EHRs) in a manner similar to the hypothetico-deductive reasoning used by physicians.

Although recent advances in EHR make it possible to discover hidden patterns from big data, most of the previous studies on EHR based analysis heavily rely on structured clinical notes and assume that the fields in such structured data have been densely filled~\cite{badnjevic2018expert,lasko2013computational,wickramasinghe2017deepr,rajkomar2018scalable, liang2019evaluation, ehwerhemuepha2018novel, aczon2017dynamic}.
However, upon admission, a large portion of patients may only have the outcomes of a small percentage of the listed possible tests, making such methods produce less than satisfactory analysis results. 
Having tests which may not be necessary so as to fill the missing fields in structured clinical notes would introduce extra burden to both hospitals and patients.
Furthermore, it is very challenging to produce fully structuralized clinical notes accurately and automatically~\cite{liang2019novel}. Consequently, there only exist limited sources of such high-quality data. 
Thus, diagnosis systems that require fully structuralized clinical notes are impractical in many real clinical scenarios.
A system that is capable of handling original unstructured clinical notes and does not alter existing admission and diagnosis procedures has great practical value.

\section{Material and Methods}

\subsection{Dataset Overview}
\begin{figure}[bt]
    \centering
    \includegraphics[width=\textwidth]{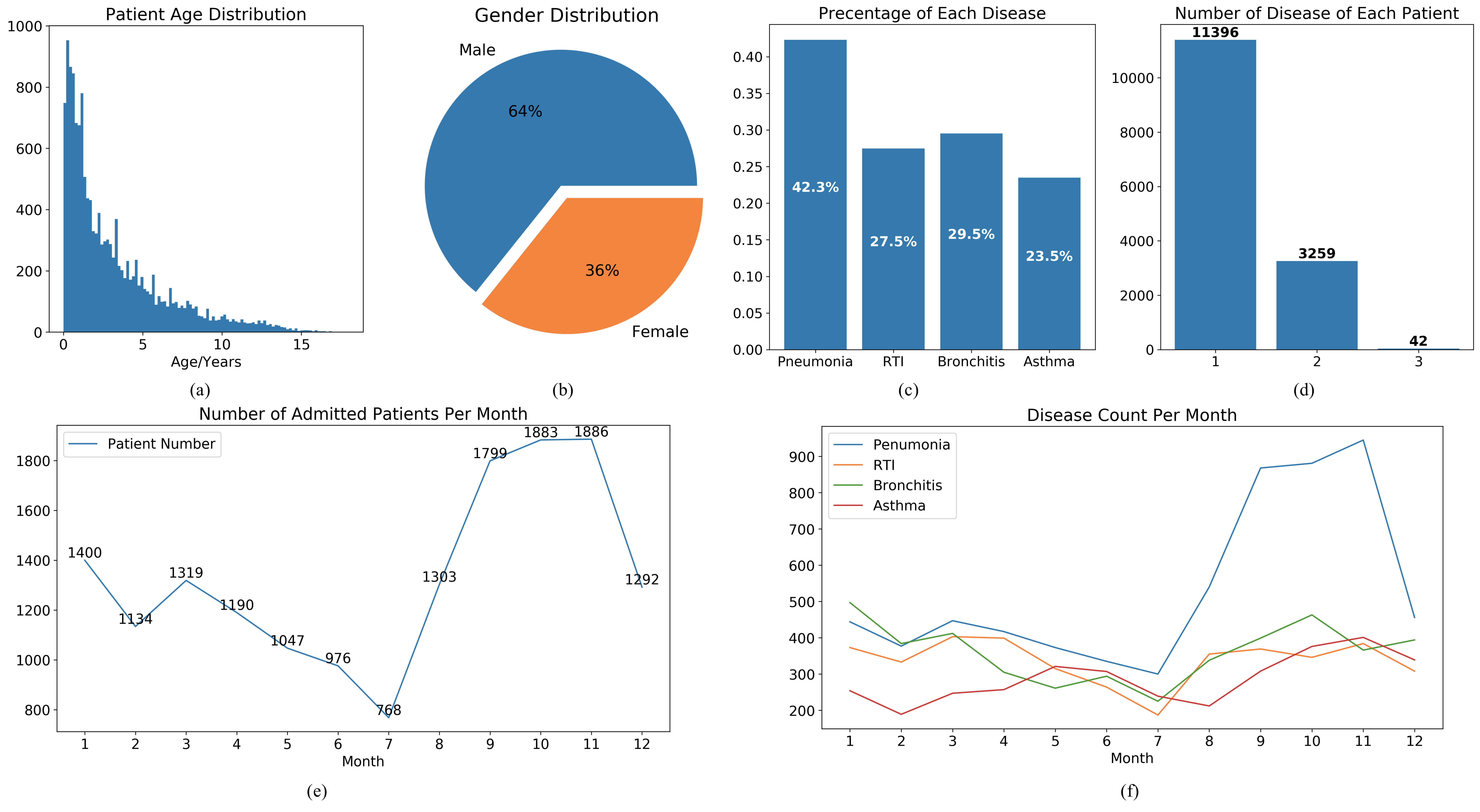} 
    \caption{Dataset statistics (better view in color version): (a) patients' age distribution, (b) patients' gender distribution, (c) patients' disease distribution, (d) distribution of disease count per patient, (e) number of admitted patients per month, (f) number of disease occurrences per month.}
    \label{fig:dataset_distribution}
\end{figure}

The admission times of all patients in the dataset range from November 2008 to July 2015.
A total of 14697 anonymous clinical notes together with their corresponding discharge diagnoses are collected from the respiratory department of the Children's Hospital of Zhejiang University School of Medicine, which is a grade AAA specialized hospital.
All clinical notes are written in Mandarin, and a few of them are translated into English and shown in the figures for easy reading.
The discharge diagnoses are used as the ground-truth disease diagnoses. 
The data is split into three non-overlapping parts.
A set of 11,100 samples is used to train models, a set of 1,821 samples is used for tuning hyperparameters and selecting the best trained model, and the rest of the 1,776 samples are used for evaluating the performance of the system. 

The ages of patients range from 1 day to 18 years old with a mean age of 3.17 and a median age of 2.00.
The detailed distribution of patients' age is shown in Fig.~\ref{fig:dataset_distribution}(a).
Over 95\% of patient ages are between 0.13 and 12 years old.
The distribution of patients' gender is shown in Fig.~\ref{fig:dataset_distribution}(b). 
64\% patients are male and 36\% are female, which is consistent with previous studies\cite{ben2018association,wang2013seasonal} and indicates higher morbidity in males.

All the patients in the dataset have at least one of the four respiratory diseases: pneumonia, RIT, bronchitis, and asthma.
The occurrence rates of these four diseases are shown in Fig.~\ref{fig:dataset_distribution}(c), and all occurrence rates vary between 23\% and 43\%.
As shown in Fig.~\ref{fig:dataset_distribution} (d), 77.5\% patients have one disease, 22.2\% patients have two of the diseases, and 0.3\% patients have three of the diseases.
The distribution of the number of diseases each patient has follows the common sense that multiple respiratory diseases were likely to occur at the same time as inflammation easily spreads in the respiratory system if proper treatment were not given in time.
Frequent occurrences of mixed symptoms from multiple diseases in the clinical notes increase the difficulty of accurate diagnosis.

We also count the number of admitted patients every month and the result is shown in Fig.~\ref{fig:dataset_distribution}(e).
The morbidity of respiratory diseases is highly related to the season. 
In autumn and winter, there are significantly more admitted patients than in summer.
This is associated with low temperature, low air humidity and poor air quality in autumn and winter. 
In Figure~\ref{fig:dataset_distribution}(f), the number of disease occurrences per month is plotted. 
The morbidity of RTI, bronchitis, and asthma have small fluctuations over the course of a year with a small increase during autumn and winter while the morbidity of pneumonia has a significant increase during autumn and winter. 
This sharp increase corresponds to the finding by S{\"a}yn{\"a}j{\"a}kangas~\cite{saynajakangas2001seasonal}.
The difference between the tendencies of pneumonia and other respiratory diseases should be attributed to the fact that during autumn and winter, respiratory diseases are likely to be aggravated and finally evolve into pneumonia. 

According to the above statistical results, the collected dataset covers a large time interval and many different real-world scenarios.
Hence, experiments conducted on this dataset can validate the effectiveness of the proposed system in clinical practice.

To simulate the clinical scenario in pediatrics, only unstructured clinical notes including chief complaints, physical examinations, and clinical test results are used to identify the diseases.
The chief complaints and physical examinations are conventional fields in the clinical notes.
In most cases, there are also results of clinical tests such as blood routine, urine routine, stool routine, and blood gas analysis.
These clinical tests can be easily conducted in primary hospitals at relatively low costs.
Moreover, since test results can be reviewed in seconds, they do not prevent a quick diagnosis.
Thus, all of this information is available upon admission and can be used by the proposed system to assist diagnosis.

\begin{figure}[tb]
    \centering
    \includegraphics[width=\textwidth]{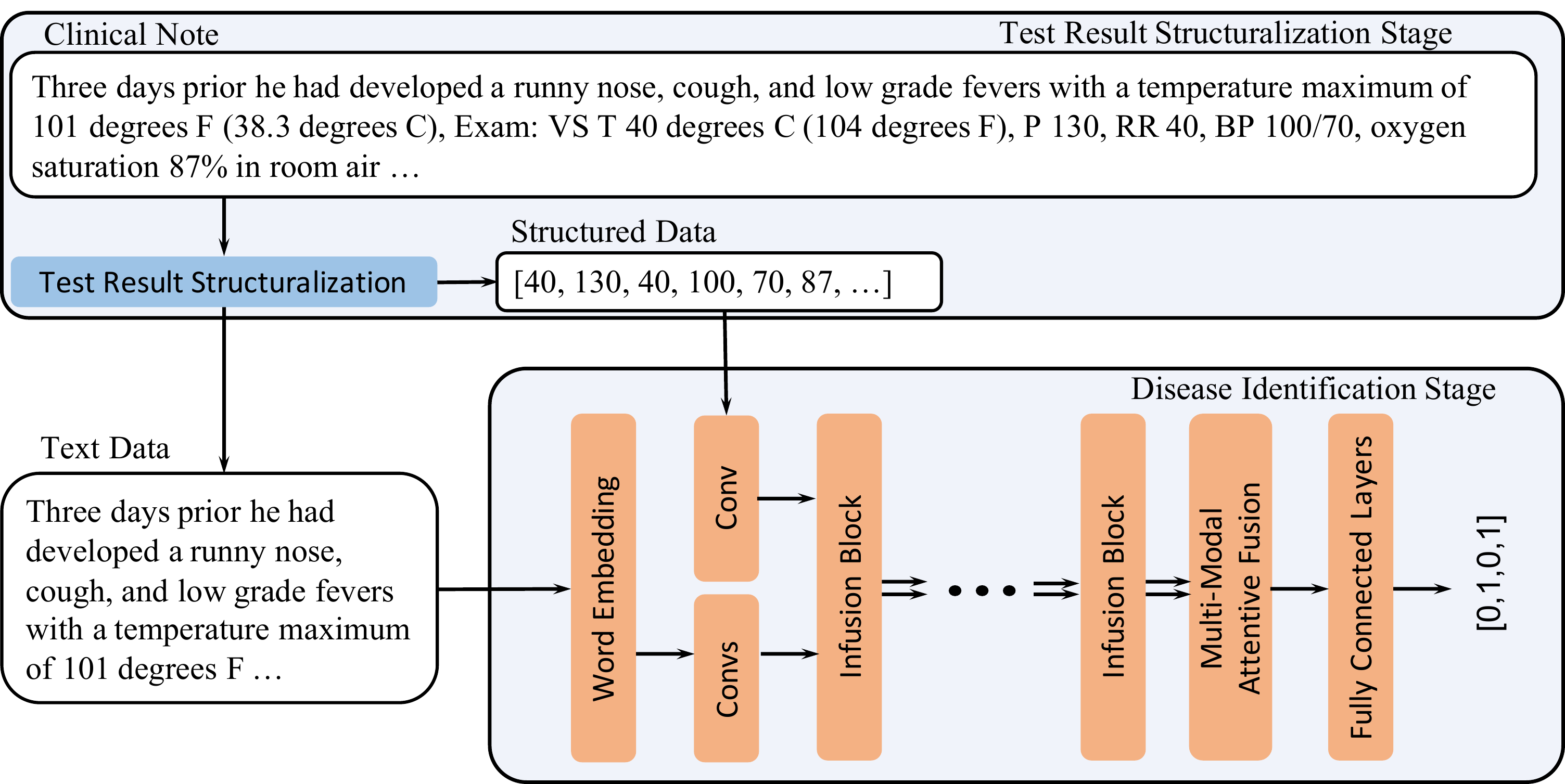}
    \caption{The overall framework of the proposed system. }
    \label{fig:framework}
\end{figure}

The overall framework of the proposed system is shown in Figure~\ref{fig:framework}.

\begin{CJK*}{UTF8}{gbsn}

\subsection{Test Result Structuralization}
The input consists of solely text from clinical notes. They are first passed into the test result structuralization stage, which extracts separate numerical test results (structured data) and text-form clinical notes (text data). Data quality is vital for system performance.
In order to make the structuralization algorithm convert the input data as cleanly as possible, a preprocessing pipeline is developed to clean up test results in raw clinical notes.
Overall, there are 7 types of pre-processing operations,
\begin{enumerate}
\item Remove physician's comments. For example, very short texts without numbers and measurements are removed, and so are plus or minus signs between brackets.
\item Remove characters for emphasizing certain parts of the text. For example, ``T (37.8 degrees C)'' is transformed to ``T 37.8 degrees C''.
\item Remove punctuations and spacing. For example, any spaces between two consecutive Chinese characters are removed to avoid misunderstanding.
\item As punctuations from different languages are often mixed up, same punctuations from different languages or different punctuations sharing the same meaning are unified. For example, the Chinese comma ``，'' and the English comma ``,'' are unified.
\item Correct obvious typos and errors. For example, the corrected version of ``T 6.5 degrees C'' is ``T 36.5 degrees C''.
\item All measurements are unified into the most commonly used version, and all numbers are transformed into scientific notations.
\item Unify the phrasing of test results. In order to make the numbers and text easier to use, we rewrite all test results, e.g. body temperature, heartbeat, blood routine, urine routine, stool routine, and blood gas, in a unified style using 201 regular expressions.
\end{enumerate}
After pre-processing, the input clinical notes are well formatted.

In order to extract and use numerical values in test results, the pre-processed text corresponding to test results in the input clinical note is further structuralized.
92 regular expressions are developed to extract the fields of commonly used test results, and meanwhile, remove the corresponding parts in the text.
There are primarily two types of regular expressions:
(1) Expressions for extracting the results of a routine blood test, e.g., ``Hemoglobin ([0-9]+[.]?[0-9]*)g/L'' for hemoglobin extraction;
(2) Expressions for extracting the results of urine routine or stool routine.
Most fields of urine and stool routine have plus or minus signs.
To fully restore the results of urine and stool routine, plus and minus signs in the results are converted into positive and negative numbers, and the regular expression for leucocyte esterase extraction can be written as ``Leucocyte Esterase [:$><$ ]*?([+\-]+|[0-9]+$\backslash$+)''.
For all fields, zero is inserted when there is no such result in the clinical note.
After the structuralization stage, every clinical note is transformed into a vector of structured data and a sequence of text data. 
In our experiments, the structured data extracted from a collection of clinical notes have very high sparsity, and the majority of fields have less than 5\% non-empty entries. To reduce the level of sparsity to a certain extent, we prune the structured data. Only fields with more than 7.5\% non-empty entries are retained.
All the retained fields in the structured data come from age, gender, vital signs, blood routine, and blood gas analysis, which define the scope of clinical tests for the proposed system. 
\end{CJK*}

\subsection{Disease Identification Stage}
The network structure of the disease identification stage is shown in the lower half of Fig.~\ref{fig:framework}.
Instead of directly processing text input, the text data is first fed into the word embedding to generate an embedding for each word.
In this paper, \textit{word2vec}~\cite{mikolov2013efficient} is employed. 
All Chinese characters are first converted into tokens by the Chinese word codebook provided by Google. These tokens are then mapped into a high-dimensional vector space $\mathcal{V}\in\mathbb{R}^C$ using the \textit{word2vec} method.
Hence, each sentence is converted into a text embedding $\bm{T}=[\bm{t}_1, \bm{t}_2, \cdots, \bm{t}_L]\in\mathbb{R}^{C\times L}$, with each $\bm{t}_i\in\mathcal{V}$.

The text embedding and structured data belong to two different domains which makes it unreasonable to merge two inputs directly.
Therefore, these two data are processed separately before being merged together. 
$\bm{T}$ is first passed into two convolution layers for initial feature extraction.
To increase the receptive field, both convolution layers have a kernel size of 3 with batch normalization and rectifier linear unit (ReLU) in between.
With these two layers, we aim to extract the information hidden beneath the adjacent words. Due to the limited length of text data we used, there are no pooling operations in these convolution layers.
The structured data $\bm{S}\in\mathbb{R}^{1\times F}$ is also passed into a convolution layer with kernel size of 1 to upsample the number of channels from $1$ to $C$.
For the sake of simplicity, in the remaining part of this paper, we use ``text stream" and ``structural stream" to indicate the networks which take text embedding and structured data as input.

\begin{figure}[tb]
    \centering
    \includegraphics[width=0.8\textwidth]{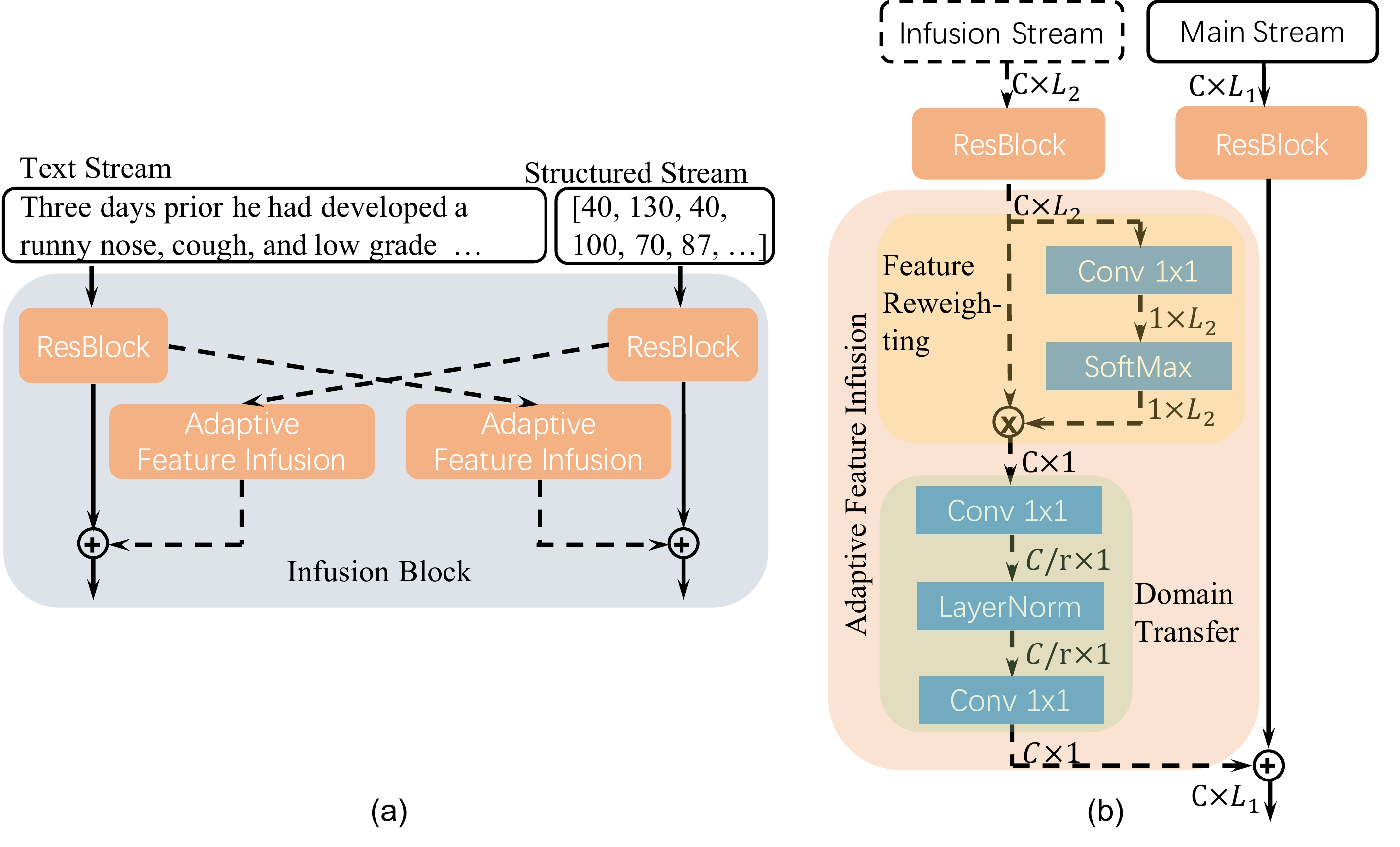}
    \caption{(a) The structure of an infusion block and (b) the detailed structure of the adaptive feature infusion module. }
    \label{fig:infusion}
\end{figure}

\begin{figure}
    \centering
    \includegraphics[width=0.3\textwidth]{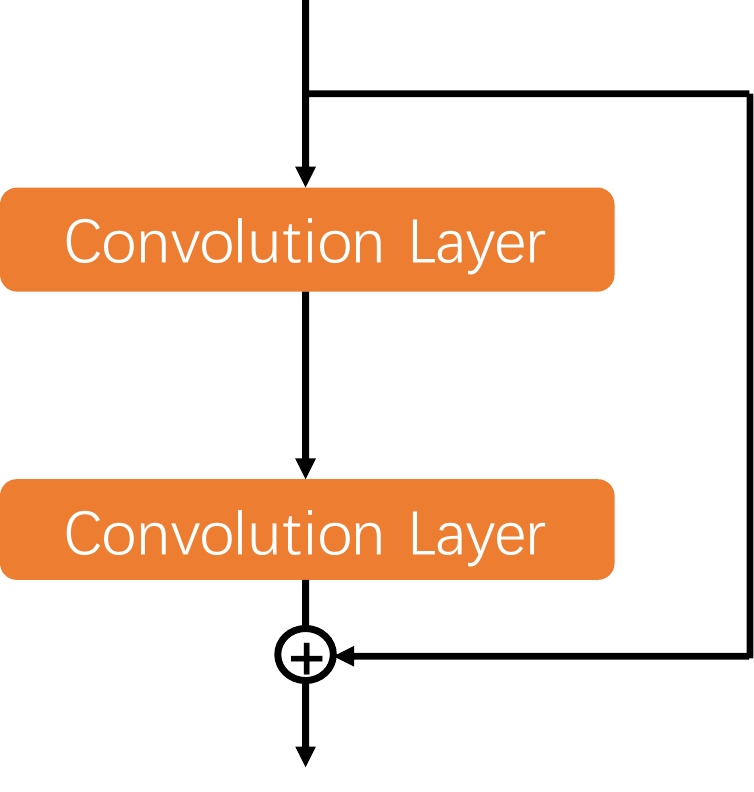}
    \caption{Structure of ResBlock}
    \label{fig:resblock}
\end{figure}

\begin{figure}[tb]
    \centering
    \includegraphics[width=0.5\textwidth]{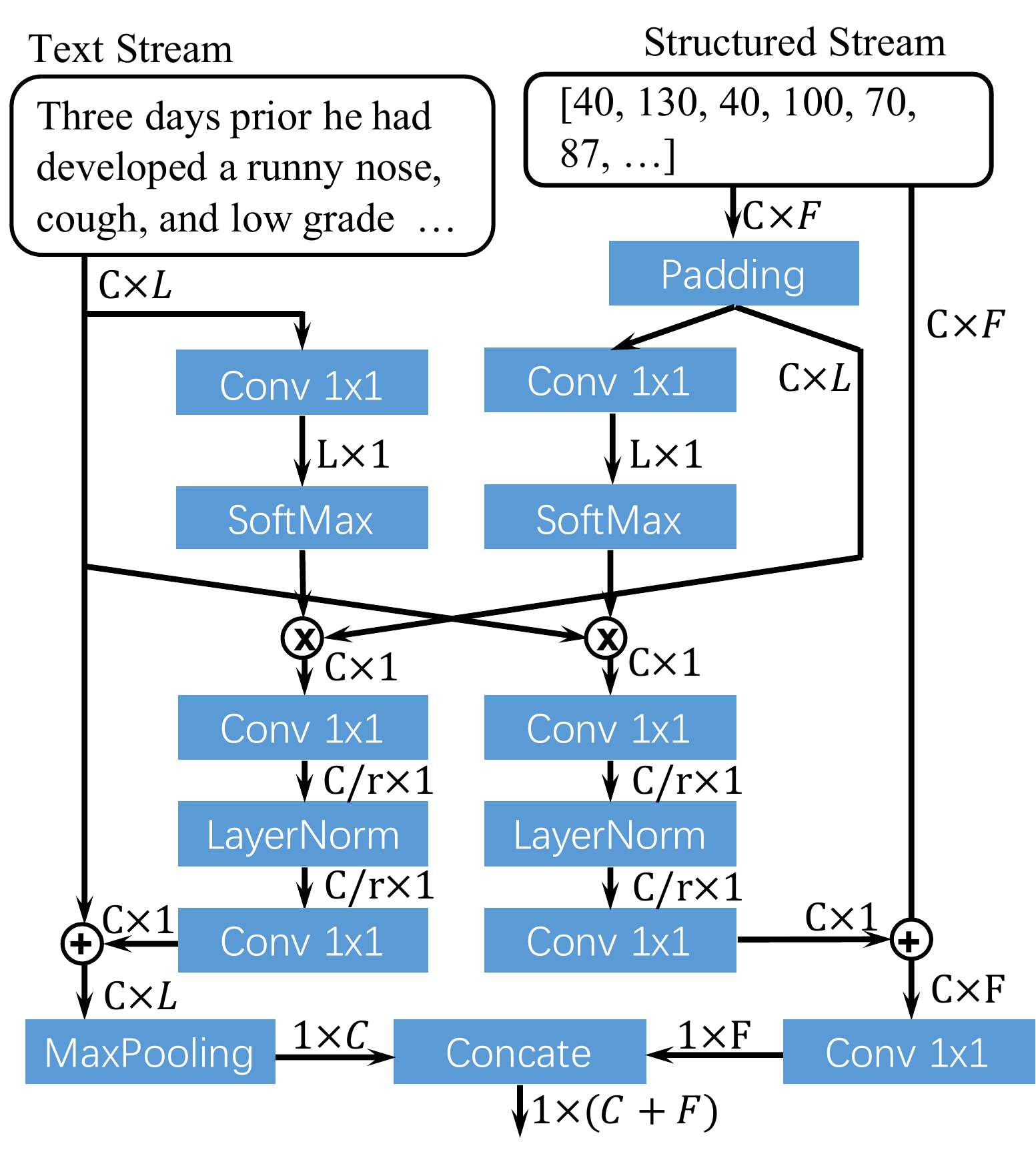}
    \caption{The structure of the multi-modal attentive fusion module. }
    \label{fig:fusions}
\end{figure}

In this paper, we focus on fine-grained disease identification.
Due to the similarity among clinical notes of fine-grained diseases, the inter-class variance is small. The identification system should find and pay more attention to discriminative phrases.
Under this guidance, we introduce an attention mechanism into the proposed system to adaptively focus on the most discriminative features in the data.
The term attention means that we additionally multiply a group of dynamic weights on the intermediate features, giving the feature that is more important to the output a larger weight, and therefore, the feature plays a more significant role in the results. This is a common concept that is widely used in contemporary machine learning works \cite{li2020behrt, devlin2018bert, wang2018non, vaswani2017attention,cao2019gcnet}.
It is well known that different depths of a network can have different observations of the input at different scales. 
In clinical notes, different observation scales correspond to different numbers of symptoms, which are related to different subsets of test results in structured data.
The numbers in structured data can be a useful supplement to the symptoms so as to understand the symptom level.
And the symptoms in text data are also able to help reveal which numbers in structured data are related.
Therefore, the infusion block is proposed to adaptively infuse the information in one stream into the other, and the multi-modal attentive fusion module is proposed to conduct dual attention between two streams and achieve feature fusion.

The outputs of two groups of convolution layers are first fed into 4 sequentially connected infusion blocks.
As shown in Fig.~\ref{fig:infusion}(a), an infusion block is a two-stream block aiming to infuse the features of $\bm{T}$ and $\bm{S}$ into each other.
$\oplus$ is the element-wise addition, which means adding each element at the same coordinate in the two matrices.
Each infusion block consists of a ResBlock and an adaptive feature infusion module for each stream.
The ResBlock is a 1-dimensional bottleneck block similar to the block proposed in ResNet~\cite{he2016deep}.
The structure of ResBlock is shown in Fig. \ref{fig:resblock}.
The structure of an adaptive feature infusion module is depicted in Fig. \ref{fig:infusion}(b) where $\otimes$ is the standard matrix multiplication operation. The order of multiplication is determined by the dimension of two input matrices.
It consists of two steps, the feature reweighting step and the domain transfer step.
For simplicity, we take the text stream as the main stream and the structural stream as the infusion stream to explain adaptive feature infusion.
The infusion stream is first passed into a convolution layer with kernel weight $\bm{W}_0$, kernel size of 1, and bias $\bm{b}_0$ to squeeze its channel number from $C$ to $1$ as
\begin{equation}
    \bm{S}_0 = \bm{W}_0 \circ \bm{S}_i + \bm{b}_0,
\end{equation} 
where $\bm{S}_i$ is the output of ResBlock in the infusion stream and $\circ$ is the convolution operation.
Softmax~\cite{gibbs1902elementary} is then applied to convert $\bm{S}_0$ into a probability distribution $\bm{\alpha}\in\mathbb{R}^{1\times F}$, with each $\alpha_{i}\in\bm{\alpha}$ follows
\begin{equation}
    \alpha_{i} = \frac{\rm{e}^{\bm{S}_{0k}}}{\sum_{j=0}^F\rm{e}^{\bm{S}_{0j}}}, 
    \label{eq:softmax}
\end{equation}
where $\bm{S}_{0k}$ represents the $k$-th element in $\bm{S}_0$, $\rm{e}$ is the base of the natural logarithm. 
$\bm{\alpha}$ is then used to reweight the elements in $\bm{S}_i$ by matrix multiplication as follows. 
\begin{equation}
    \bm{S}_r = \bm{S}_i\bm{\alpha}^T.
\end{equation}
Thus, elements in $\bm{S}_i$ containing more discriminative features can have higher weights, increasing the inter-class variance and boosting the performance of the system in identifying fine-grained samples.
As $\bm{S}_r$ can be considered as a weighted version of $\bm{S}_i$, both $\bm{S}_i$ and $\bm{S}_r$ belong to the domain of structured data.
However, with significant differences between the text domain and the structured data domain, directly fusing data from one domain into the other does not make sense.
As discovered in CycleGAN~\cite{zhu2017unpaired}, data samples from one domain can be effectively converted into another domain with the essential information preserved using an encoder network. 
Thus, two convolution layers with a layer normalization layer in between are used to transfer the structured stream into the text stream in the domain transfer step.
The size of both convolution kernels are set to 1 and the number of channels in between is set to $\frac{C}{r}$ (where $r$ is 4).
The output of domain transfer is then replicated $L$ times before being fused element-wise into the text stream.

The structure of the multi-modal attentive fusion module is shown in Fig. \ref{fig:fusions}, where the text stream and structural stream are thoroughly fused.
As the length $F$ of the structured data is shorter than the length $L$ of the text embedding, the matrix $\bm{S}_m$ in the structural stream is zero padded to the same length as the text embedding.
For the sake of simplicity, we refer to the padded structured data also as $\bm{S}_m$.
$\bm{S}_m\in\mathbb{R}^{C\times L}$ and the matrix $\bm{T}_m\in\mathbb{R}^{C\times L}$ in the text stream are then sequentially passed into their corresponding convolution layer with the kernel size equal to 1 and the softmax layer, becoming probability distributions $\bm{S}_p\in\mathbb{R}^{1\times L}$ and $\bm{T}_p\in\mathbb{R}^{1\times L}$, respectively. 
Inspired by non-local neural networks~\cite{wang2018non} and GCNet~\cite{cao2019gcnet}, extracting pairwise relationships with matrix multiplications is a productive approach to merge information. 
Therefore, $\bm{S}_p$ and $\bm{T}_p$ are used to reweight $\bm{T}_m$ and $\bm{S}_m$ through matrix multiplications, attentively optimizing the information in $\bm{T}_m$ and $\bm{S}_m$ into $\bm{T}_w$ and $\bm{S}_w$, respectively, and augmenting the features that best identify the diseases.
Similar to adaptive feature infusion, $\bm{T}_w$ and $\bm{S}_w$ are fed into encoder networks to achieve domain transfer followed by elementwise addition to fuse the two streams.
The output vectors of domain transfer are replicated $L$ and $C$ times in the text stream and structural stream respectively before addition. 
Take the structural stream as an example, the above process can be formulated as follows,
\begin{equation}
    \bm{S}_o = \bm{S}_m + \bm{W}_d\circ(\text{LN}(\bm{W}_e\circ(\text{SoftMax}(\bm{W}_s\circ\bm{S}_m+\bm{b}_s)\bm{T}_m^T)+\bm{b}_e))+\bm{b}_d,
\end{equation}
where $\bm{S}_o$ and $\bm{T}_o$ are the outputs of the multi-modal attentive fusion module in the structural and text streams, respectively, $\bm{W}_s,\ \bm{W}_e,\ \bm{W}_d$ and $\bm{b}_s,\ \bm{b}_e,\ \bm{b}_d$ are the kernel weight matrices and bias vectors of convolution layers, and LN and SoftMax represent the layer normalization~\cite{ba2016layer} and softmax operations~\cite{gibbs1902elementary}. 
According to the properties of fine-grained identification, for the text stream, max pooling with transpose converts $\bm{T}_o\in\mathbb{R}^{C\times L}$ into $\bm{T}_c\in\mathbb{R}^{1\times C}$ so as to keep only the most discriminative feature.
For the structural stream, a convolution layer is used to merge all channels and convert $\bm{S}_o\in\mathbb{R}^{C\times F}$ into $\bm{S}_c\in\mathbb{R}^{1\times F}$.
To fully fuse the information in these two streams, $\bm{T}_c$ and $\bm{S}_c$ are concatenated to form $\bm{M}\in\mathbb{R}^{1\times (C+F)}$, which is the output of the multi-modal attentive fusion module.

Finally, $\bm{M}$ is passed into a dropout layer and two fully-connected layers with a ReLU in between, generating the output $\bm{P}_o\in\mathbb{R}^{4}$ of the proposed system.
Each value in $\bm{P}_o$ represents the prediction score of one of the target respiratory diseases.
Binary Cross-Entropy Loss is used to train the proposed system.

\subsection{Performance Metrics}
The proposed system considers the identification problem of the four diseases as a multi-label classification task.
As the disease sample distribution is imbalanced, the accuracy cannot comprehensively evaluate the overall performance of the system.
In this paper, we use mean average precision (mAP) as the main performance metric.
Average precision (AP) is the area under the precision-recall curve of one category,
\begin{equation}
    \text{AP} = \int_0^1 p(r)dr,
\end{equation}
where $p$ and $r$ are precision and recall, respectively.
They are defined as 
\begin{equation}
\begin{aligned}
    p &= \frac{\textbf{TP}}{\textbf{TP}+\textbf{FP}} \\
    r &= \frac{\textbf{TP}}{\textbf{TP}+\textbf{FN}}, 
\end{aligned}
\end{equation}
where $\textbf{TP},\ \textbf{FN},\ \textbf{FP}$ are true positive rate, false negative rate and false positive rate, respectively, and mAP is then calculated as the average value of APs over four diseases. 

As $F_1$ score is used to evaluate the performance in several related works, we also report $F_1$ score as an auxiliary performance metric.
$F_1$ score is defined as follows:
\begin{equation}
    F_1 = 2\times \frac{p\times r}{p + r}. 
\end{equation}
In this paper, we report the marco $F_1$ score, which is the unweighted mean $F_1$ score of each disease. All precisions (p) and recalls (r) are computed under a fixed threshold of 0.5 (applied to the sigmoid outputs of $\bm{P}_o$).

To provide an in-depth understanding of the performance, a confusion matrix is also used to visualize the performance. 
We report the confusion matrix in the format as in Table \ref{tab:confusion matrix format}.
\begin{table}[]
    \centering
    \begin{tabular}{c|c}
        TN & FP \\
        \midrule
        FN & TP
    \end{tabular}
    \caption{Format of confusion matrix we use. }
    \label{tab:confusion matrix format}
\end{table}

To evaluate the significance of any improvement, $p$-value is used as another measurement.
Following the evaluation protocol in a recent \textit{Nature} article~\cite{mckinney2020international}, the system is first evaluated on the test dataset using the Bootstrap~\cite{johnson2001introduction} based resampling method.
And a $p$-value on the difference between the baseline and proposed method was generated through the use of a permutation test~\cite{chihara2011mathematical}.
The reported two-sided $p$-value is obtained by comparing the observed statistic to the empirical quantiles of the randomization distribution.

To find out the relation between diseases and test result fields, their correlations are reported and defined as
\begin{equation}
    \text{Cov}(x, y) =  \frac{\max \limits_{k}(\sum_{i=1}^N x[i+k] * y[i])}{N},
\end{equation}
where $x$ and $y$ are one field of structured data (samples which have no such field is ignored) and one disease in the test set, respectively.

\section{Results}
\subsection{Disease Identification Performance}

\begin{table}[t]
    \centering
    \resizebox{\columnwidth}{!}{
    \begin{tabular}{cccccccccc}
        \toprule[1.5pt]
        Method & Pneumonia & RTI & Bronchitis & Asthma & Acc & mAP & $F_1$ & P & R \\
        \hline 
        SVM & 0.679 & 0.670 & 0.429 & 0.586 & 0.735 & 0.591 & 0.406 & 0.409 & 0.413\\
        Decision Tree & 0.589 & 0.506 & 0.377 & 0.460 & 0.733 & 0.482 & 0.549 & 0.552 & 0.556 \\
        Logistic Regression & 0.668 & 0.647 & 0.434 & 0.569 & 0.729 & 0.579 & 0.422 & 0.424 & 0.438\\
        \hline
        WordCNN \cite{kim2014convolutional} & 0.838 & 0.817 & 0.609 & 0.648 & 0.816 & 0.728 & 0.671 & 0.677 & 0.675\\
        ConvNN & 0.818 & 0.815 & 0.637 & 0.768 & 0.823 & 0.759 & 0.702 & 0.710 & 0.707 \\
        BLSTM \cite{graves2005framewise} & 0.806 & 0.762 & 0.570 & 0.680 & 0.814 & 0.704 & 0.681 & 0.683 & 0.689\\
        Covid19NN \cite{obeid2020artificial} & 0.892 & 0.791 & 0.647 & 0.787 & 0.832 & 0.779 & 0.683 & 0.687 & 0.688 \\
        \hline
        BERT \cite{devlin2018bert} & 0.844 & 0.772 & 0.569 & 0.672 & 0.836 & 0.714 & 0.673 & 0.674 & 0.682 \\
        LMF \cite{liu2018efficient} & 0.830 & 0.782 & 0.573 & 0.655 & 0.806 & 0.709 & 0.653 & 0.657 & 0.661 \\
        \hline
        \textbf{Proposed} & \textbf{0.878} & \textbf{0.857} & \textbf{0.714} & \textbf{0.825} & \textbf{0.845} & \textbf{0.819} & \textbf{0.742} & \textbf{0.750}& \textbf{0.752}\\
        \bottomrule[1.5pt]
    \end{tabular}
    }
    \caption{Performance comparison between the proposed system and multiple benchmark algorithms. }
    \label{tab:main}
\end{table}

Identification results of the proposed system and a variety of baseline methods are reported in Table~\ref{tab:main}. 
To guarantee the fair comparison, all baseline methods are evaluated with the same structuralized data as the proposed system and are trained on the same training dataset.
As the dataset is imbalanced, aside from the average accuracy on each disease, we also report the AP of each disease, mAP, average $F_1$ score, precision (P), and recall (R) to provide a comprehensive evaluation of the proposed system.
It is worth noting that the reported $F_1$ score, P, and R are calculated by the mean value of disease-wise $F_1$ score, P, and R, respectively. Thus, the relationship of $F_1=\frac{2PR}{P+R}$ is not strictly satisfied in the reported results.
Traditional machine learning based methods, including support vector machine (SVM), decision tree and logistic regression, and a variety of deep-learning-based methods recently used in clinical note analysis, including word embedding based CNN for Covid-19 (Covid19NN) \cite{obeid2020artificial}, WordCNN~\cite{kim2014convolutional}, BLSTM~\cite{graves2005framewise}, and a two-layer plain CNN (ConvNN), are used for comparison. 
Three traditional machine learning based methods have been implemented using the Scikit Learn package~\cite{scikit-learn} in Python and fine-tuned with the \textit{GridSearchCV} function in Scikit Learn by exhaustively trying different parameter combinations and selecting the best one. 
The deep-learning-based methods have been tested using the authors' original open source codes. 
It is interesting that the proposed system significantly outperforms all the other methods by a large margin on all performance metrics.
According to Table~\ref{tab:main}, the deep-learning-based methods improve the mAP from below 0.6 to about 0.75, and our proposed system further pushes the mAP to 0.819, reaching a new level of performance.

The Covid19NN proposed for predicting Covid-19 based on EHR records represents the most recent advance in EHR analysis. 
In their network, they concatenated three convolution layers with different kernel sizes, forming a block structure like the naive version Inception module proposed in GoogLeNet \cite{szegedy2015going}. 
Then, one embedding layer, one Inception module-like block, and two fully-connected layers are sequentially connected to construct the Covid19NN. 
Compared with their approach, our proposed system has a significantly higher performance. 
Specifically, the proposed system achieves 0.012, 0.04, and 0.059 higher in Accuracy, mAP, and $F_1$ score, respectively. 
BLSTM~\cite{shickel2017deep, chalapathy2016bidirectional,liu2018deep, wunnava2018bidirectional} and plain CNN (ConvNN)~\cite{liu2018deep, suresh2017clinical} are the deep learning architectures most frequently used in the state-of-the-art clinical note analysis methods. 
WordCNN~\cite{kim2014convolutional} has been proved to be a light-weight but effective architecture in multiple text analysis tasks. 
Thus, we compare the proposed system with these methods. 
In comparison to WordCNN, ConvNN and BLSTM, the proposed system achieves 0.091, 0.060, and 0.115 improvement on mAP and 0.071, 0.040, and 0.061 on $F_1$ score, respectively.
The outstanding performance of the proposed system primarily benefits from the adoption of ResBlock, adaptive feature infusion, and multi-modal attentive fusion.
Both WordCNN and ConvNN use similar network architectures and share the difficulty of training deeper networks.
All of these three existing networks cannot handle numbers properly and end up losing the important numerical information in clinical tests.
Another important aspect of the proposed system is its ability in extracting fine-grained features from a vast amount of similar data.
By amplifying minor differences among different clinical notes, the proposed system is able to classify every note into different diseases.

To better validate the proposed system, we also compare our system with the state-of-the-art natural language processing model BERT \cite{devlin2018bert} and multi-modality fusion model LMF \cite{liu2018efficient}. 
BERT is a massive model designed for multiple tasks including sequence classification, while LMF is a low-rank multi-modal data fusion model.
In comparison to BERT and LMF, the proposed system still achieves significant improvements in performance. 
The mAP is improved by 0.105 and 0.110, and the $F_1$ score is improved by 0.069 and 0.089 in comparison to BERT and LMF, respectively.
The inferior performance of BERT is attributed to the following two factors.
First, the language used in clinical notes is relatively simple. 
The most significant advantage of BERT over other models is its ability to analyze complex language. 
This advantage does not match the low complexity of clinical notes. 
Specifically, it is a common practice that physicians copy/paste or use templates when taking notes, which leads to the simplicity of clinical note structure.
Thus, language complexity is not the performance bottleneck of the system.
Second, massive models like BERT usually need to be pre-trained on a gigantic dataset.
However, all the existing BERT models are pre-trained on general purpose language datasets, which is not suitable for the analysis of clinical notes. 
Thus, using BERT in this scenario may even hurt the performance. 
The superior performance of the proposed system over LMF is attributed to its ability in effectively exploiting fine-grained features.
As LMF adopts low-rank properties to fuse multi-modal information, details crucial for fine-grained identification are likely to be omitted.

\begin{figure}
    \centering
    \includegraphics[width=\textwidth]{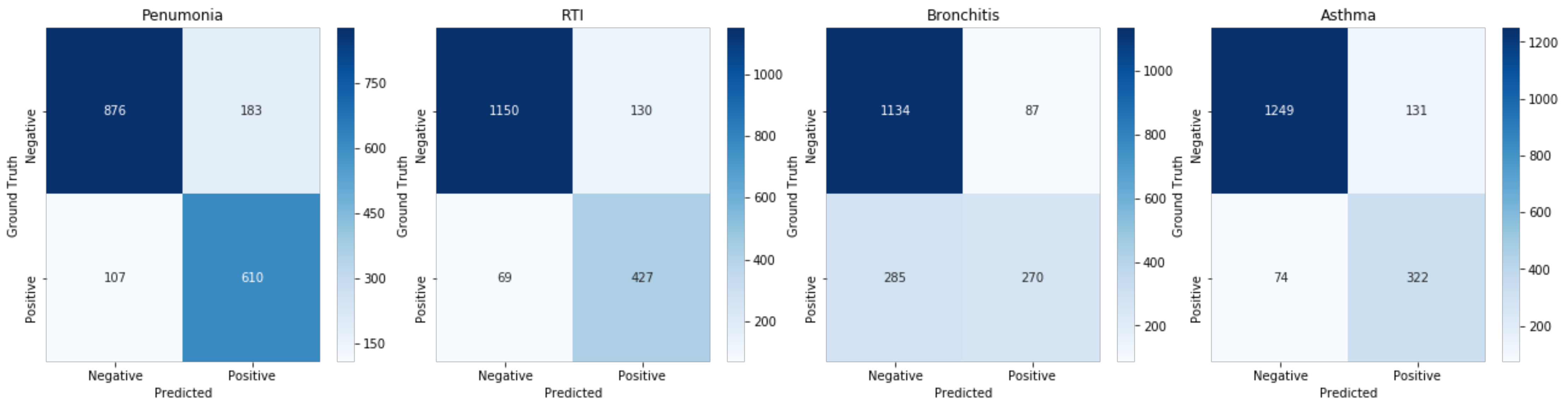}
    \caption{Confusion matrices of each studied disease. }
    \label{fig:disease confusion matrix}
\end{figure}

To provide an in-depth understanding of the proposed system's results, we further report the confusion matrix of each disease in Fig. \ref{fig:disease confusion matrix}. 
From the figure, it can be clearly seen that the proposed system achieves impressive overall performance. The FP and FN value on pneumonia, RTI and asthma are limited to a reasonable range. And the system's bias to detecting a specific disease is not significant, which makes the FP and FN value distributed evenly. 
However, it is necessary to point out that bronchitis suffers from a slightly high FN value. 
We blame such performance drop to the atypical symptom of bronchitis. 
The identification of bronchitis in pediatrics with limited information from clinical notes should be further studied in future works.

\subsection{Ablation Study}
%The results are shown in Table \ref{tab:data},\ref{tab:arch_depth},\ref{tab:fusion}. 
%In the table, Text and structural represents using original clinical notes and structural data as input to the CNN, respectively. 
%ConvNN represents the word embedding with convolution layers and fully connected layers in the disease identification stage, which is similar to the artificial neural network architecture adopted in most of the clinical note analysis systems \cite{badnjevic2018expert,liu2018deep, miotto2016deep,craig2017predicting,rattanajariya2019applying}. 
%ResNet represents the proposed system without the adaptive feature infusion and multi-modal attentive fusion. 
%In the ConvNN and ResNet with both text and structural data as input, only concatenation is used to fuse the multi-modal data. 
%The benchmark column indicates the comparison method used to evaluate if the performance improvement is significant. 
\subsubsection{Effectiveness of Multi-Modal Data}\label{subsubsec:multi_modal}

\begin{table}[t]
    \centering
    \begin{tabular}{ccccc}
        \toprule[1.5pt]
       Data  & Architecture & Benchmark & mAP & $p$-value \\
        \hline
        Raw & ConvNN & - & 0.746 & - \\
        Text & ConvNN &  - & 0.745 & - \\
        structured & ConvNN & - & 0.524 & - \\
        \hline
        Text + structured & ConvNN & Text + ConvNN & 0.759 & $p$<0.01 \\
        Text + structured & ConvNN & structured + ConvNN & 0.759 & $p$<0.001 \\
        Text + structured & ConvNN & Raw + ConvNN & 0.759 & $p$<0.01 \\
        \bottomrule[1.5pt]
    \end{tabular}
    \caption{A comparison of performance associated with different inputs.}
    \label{tab:data}
\end{table}

In order to prove the effectiveness of multi-modal input, some experiments were conducted, and the results are shown in Table~\ref{tab:data} (``Raw'' indicates the original clinical notes without any modification, ``Text'' and ``structured'' indicate the text data and structured data).
To evaluate the performance without the influence of network structure, ConvNN, which has the same architecture as the convolution layers in the text stream, is used. The text feature is fused with structured data before fully-connected layers.
We report mAP as the performance metric and $p$-value to verify the significance.
When using only text data or structured data, the mAP is 0.745 or 0.524, respectively.
The performance of structured data is far below that of text data.
This is because only blood routine and blood gas are extracted as the structured data.
%If more structural fields are available, structural data can be as effective as text data.
With the raw data, ConvNN achieves a mAP of 0.746, which is almost the same as the mAP associated with the text data.
The similar performance of raw data and text data indicates that, by using test results as text, the algorithm cannot understand the meaning of the numbers hence no improvement is achieved.
However, the system enjoys an improvement of 0.014, 0.235, and 0.013 in mAP respectively by naively concatenating the text data and structured data before passing them into the fully-connected layers.
We compute $p$-values for the comparisons between the performance associated with multi-modal input and the performance associated with the three types of single-modal inputs, and the small $p$-values verify the statistical significance of the performance improvement achieved with multi-modal input.
Though structured data is extracted from clinical notes with high sparsity, using structured data as independent input can help the system distinguish different fields easily so that the system can compare the same field from different notes to learn the difference between normal and abnormal values.
The significant performance improvement achieved with multi-modal input over single-modal inputs verifies the superiority of multi-modal data.
In the rest of the paper, we will also verify that, by utilizing multi-modal input in a well-designed network, the potential of multi-modal data can be further explored.

\subsubsection{Backbone Block Architecture}\label{sec:arch}

\begin{table}[t]
    \centering
    \begin{tabular}{cccccc}
    \toprule[1.5pt]
       Backbone & \# Blocks & mAP & $\Delta$ mAP\\
        \hline
        ConvNN & - & 0.759 & - \\
        \hline
        ConvNN + PlainConv & 2 & 0.766 & 0.007\\
        ConvNN + PlainBlock & 2 & 0.793 & 0.034\\
        ConvNN + ResBlock & 2 & 0.795 & 0.036 \\
        \hline
        ConvNN + PlainConv & 4 & 0.753 & -0.006 \\
        ConvNN + PlainBlock & 4 & 0.775 & 0.016 \\
        ConvNN + ResBlock & 4 & 0.799 & 0.040 \\
        \hline
        ConvNN + ResBlock & 8 & 0.800 & 0.041 \\
        \bottomrule[1.5pt]
    \end{tabular}
    \caption{Performance comparison among different backbone network architectures and depths. }
    \label{tab:arch_depth}
\end{table}

In most recent works on clinical notes and EHR analysis, plain convolutional architectures are widely adopted~\cite{badnjevic2018expert,liu2018deep, miotto2016deep,craig2017predicting,rattanajariya2019applying}.
The assumption is that plain CNNs are powerful enough to fit the hidden patterns in the notes.
However, inspired by the works in Computer Vision~\cite{he2016deep}, we suggest that the architecture of the backbone network is also vital for the performance of disease identification systems.
To verify the effect of different architectures, we compare plain convolution (PlainConv) with ResBlock. 
For fair comparison, each PlainConv consists of two convolution layers, which has a similar complexity with ResBlock. 
We also test PlainBlock, which is a ResBlock without the identity mapping branch, as a more complicated version of plain CNNs.
The experimental results are shown in Table~\ref{tab:arch_depth}.
ConvNN is the 2-layer CNN in Table~\ref{tab:data}. 
``\# Blocks'' indicates the number of PlainConv/PlainBlock/ResBlock used in the network.
``$\Delta$ mAP'' is the improvement in mAP compared with baseline ConvNN.
For PlainConv and PlainBlock, as the network depth grows, the mAP first increases then decreases.
The performance of ResBlock keeps improving with more than 8 blocks. 
And in all settings, ResBlock is able to outperform the other two blocks.
The outstanding performance of ResBlock proves that the network architecture can affect the system performance greatly.
Even though more blocks result in better performance, the great computational complexity makes it inappropriate to be applied in practice.
Hence, in this paper, 4 ResBlocks are used as the backbone architecture.

%can express a function that is complex enough to fit the complexity of the clinical data. 
%However, comparison between the performance of ConvNN and ResNet demonstrates that by properly increasing the complexity of the model, the performance of the model can be significantly improved. 
%The increased 
\subsubsection{Feature Fusion}

\begin{table}[t]
    \centering
    \begin{tabular}{ccccc}
    \toprule
        Architecture & mAP & $\Delta$ mAP & $p$-value\\
        \hline
        ConvNN & 0.759 & - & -\\
        ResBlock & 0.799 & 0.040 & $p$ < 0.001 \\
        Infusion Block & 0.813 & 0.054 & $p$ < 0.001\\
        Infusion Block + Fusion & 0.819 & 0.060 & $p$ < 0.001\\
    \bottomrule
    \end{tabular}
    \caption{The performance contribution of different module combinations. The benchmark method used for comparison is ConvNN.}
    \label{tab:fusion}
\end{table}

To evaluate the contributions of adaptive feature infusion, multi-modal attentive fusion and their combination, an ablation study was conducted and the results are reported in Table~\ref{tab:fusion}.
In these experiments, ``ResBlock'' indicates ``ConvNN+ResBlock-4'' in Table~\ref{tab:arch_depth}, ``Infusion Block'' indicates the system in Fig.~\ref{fig:framework} without multi-modal attentive fusion and ``Infusion Block+Fusion'' indicates the whole system.
By replacing ResBlock with Infusion Block, the mAP increases from 0.799 to 0.813 and has a 0.054 improvement over ConvNN.
Multi-modal attentive fusion further pushes the mAP to 0.819, which is 0.006 higher than that of Infusion Block.
All methods are also used to compare with ConvNN and the resulting very small $p$-values support the significance of these methods.
The significant performance improvements of the proposed components are attributed to their ability in attentively fusing information from both streams.
Normally, symptoms and clinical test results are strongly correlated and reveal each other.
In reality, doctors conduct pre-diagnosis on the basis of their observation of the symptoms and look for evidence in the test results.
Infusion block and multi-modal attentive fusion imitate these steps and the prediction is based on both streams.
The proposed method also benefits from attentive fusion, which reweighs similar features and amplifies the differences between fine-grained clinical notes.

\subsubsection{Symmetric Model Design}
As the data in the text and structural streams are quite different, the model depth needed to exploit the information hidden beneath each stream may differ. 
Thus, we further test if asymmetric model design can utilize such differences to achieve superior performance. 
We evaluate the mAP of asymmetric model designs with different depths for different data streams.
Results are shown in Table \ref{tab:asym}, where the model depth for the text and structural streams represents the number of infusion blocks used in each stream, respectively. 
The model with four blocks for both text and structural streams is the symmetric model design used in the proposed system. 
As shown in the table, the asymmetric models do not achieve significant differences in performance.
This indicates that, although the text and structural data may be quite different, the main reason for the performance is the effective fusion of multi-modal data but not the network depth for each stream.

\begin{table}[t]
    \centering
  		\begin{tabular}{ccc}
  			\toprule
  			\multicolumn{2}{c}{Depth} & \multirow{2}*{mAP}\\
  			\cline{1-2}
  			Text & Structural & ~ \\
  			\hline
  			2 & 4 & 0.815 \\
  			4 & 2 & 0.814 \\ 
  			$\textbf{4}$ & $\textbf{4}$ & $\textbf{0.819}$ \\
  			8 & 4 & 0.816 \\
  			4 & 8 & 0.815 \\
  			\bottomrule
  		\end{tabular}
    \caption{Performance comparison between symmetric and asymmetric model designs.}
    \label{tab:asym}
\end{table}

%In the adaptive feature infusion, the information in one domain is used to calculate an attention vector, helping the system to decide which part is more impactful in the other domain. 
%Specifically, as the CNN focuses on different levels of feature in different depth, the adaptive feature infusion in different infusion blocks helps the system to look at different kind of features better.
%Thus, the system understands the data more thoroughly. 
%While in the multi-modal attentive fusion, the data in one domain is added to the other domain with attentive modulation of data from the other domain before concatenation. 
%As the text and structural data are from different domains and processed separately, this procedure better matches data before concatenation. 

%\subsubsection*{Multi-Label Learning}
%\begin{table}[t]
%    \centering
%    \begin{tabular}{c|cccc| c|c}
%       Method &  Pneumonia & RTI & Bronchitis & Asthma & mAP & $F_1$ \\ 
%       Single-Label & 0.866 & 0.829 & 0.632 & 0.820 & 0.786 & 0.730 \\
%       Multi-Label  & 
%    \end{tabular}
%    \caption{The Performance comparison of the multi-label learning and }
%    \label{tab:my_label}
%\end{table}
%

\subsection{Interpretability}

\begin{figure}
    \centering
    \includegraphics[width=1.0\textwidth]{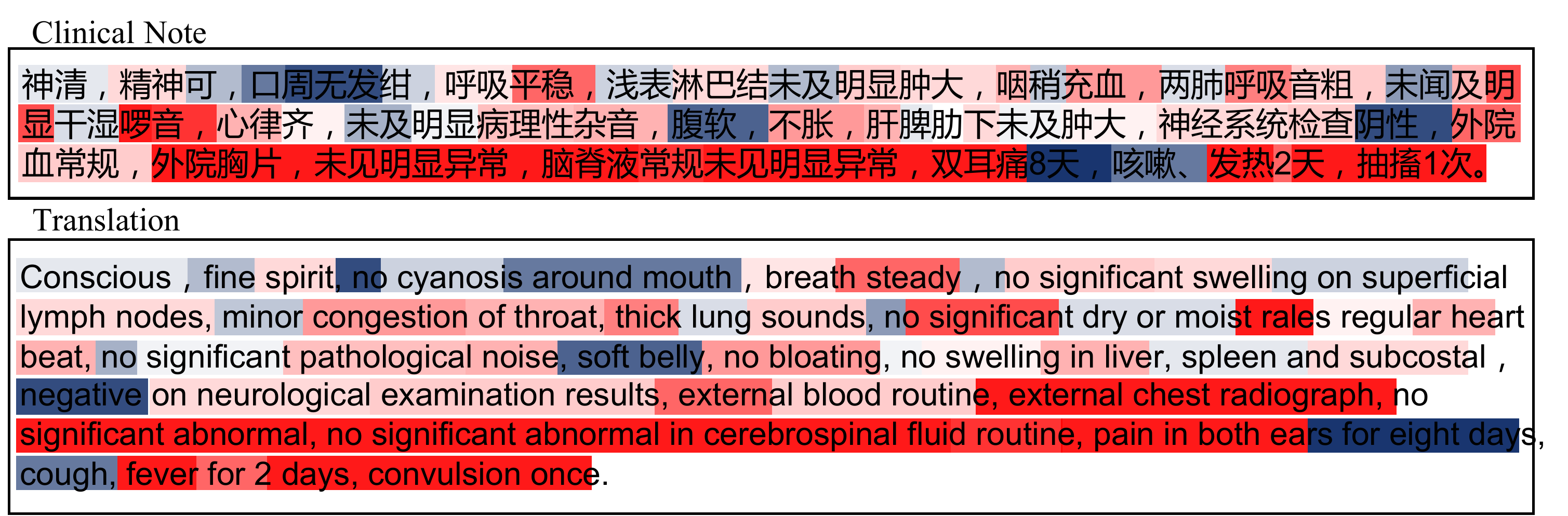}
    \caption{The gradient magnitude of the identification output for pneumonia with respect to word embedding. The upper half is the original text data in Chinese and the lower half is the translated version.}
    \label{fig:gradient}
\end{figure}

The ability to provide interpretable results is the key to understand why and how the trained model works.
The interpretability can also help doctors decide when to accept a prediction and when to ignore an unexpected result.
As shown in Fig.~\ref{fig:gradient}, we visualize the gradient magnitude of the identification output with respect to each word to depict how each word influences the identification output.
A blue color represents a negative gradient while a red color represents a positive gradient.
The darker a color is, the larger gradient the word has.
These different colors can help understand inner-phrase relationships and locate crucial information to diagnose diseases.

Take the phrase``\textit{negative on neurological examination results}'' as an example.
The gradient with respect to ``\textit{neurological examination}'' is relatively small as this is a common clinical test.
The word ``\textit{negative}'' is a strong indicator showing that the patient appears to be normal in this test, thus, a significant negative gradient is associated with this word.
A similar inner-phrase contrast appears in multiple areas, such as ``\textit{no significant swelling on superficial lymph nodes}'', ``\textit{minor congestion of throat}'', ``\textit{no significant dry or moist rales}'' and ``\textit{no significant pathological noise}''.
This behavior flips the meaning of a phrase with negative words.

The system's ability in locating crucial information is supported by the gradient distribution.
In the first half of the clinical note, the text data consists of physical examination results, which is similar among patients.
Thus, the average gradient magnitude is relatively small, and the system only focuses on a few key words.
While in the second half of the note, the text data consists of previous tests and the chief complaint of the patient.
These two parts are the key to diagnosis, as the chief complaint shows the most important symptoms of a patient and the past tests indicate diseases the patient may have.
The words in the second half are associated with gradients with significantly higher magnitude, showing that our system automatically learns to use crucial information as the main basis of diagnosis.

\begin{table}[]
    \centering
    \resizebox{\columnwidth}{!}{
    \begin{tabular}{c|cccccccccc}
    \toprule[1.5pt]
        Disaese & Gender & Age & Temp. & HR & BR & SBP & DBP & SPO2 &WBC & N \\
        \hline
        Pneumonia & 0.144 & 0.069 & 0.041 & 0.218 & 0.033 & 0.001& 0.035 & 0.032 & 0.010& 0.230  \\
        RTI & 0.107 & 0.075 & 0.026& 0.126 & 0.016 & 0.001 & 0.026 & 0.003 & 0.007 & 0.153 \\
        Bronchitis & 0.103 & 0.043 & 0.034 & 0.173 & 0.025 & 0.001 & 0.030 & 0.017 & 0.007 & 0.155 \\
        Asthma& 0.064 & 0.048 & 0.024 & 0.128 & 0.019 & 0.001 & 0.023 & 0.021 & 0.005 & 0.144 \\
        \toprule[1.5pt]
          & CRP & HGB & PLT & LY & BPH & PCO2 & PO2 & K & Na \\
          \hline
        Pneumonia & 0.016 & 0.236 & 0.120 & 0.026 & 0.465 & 0.226 & 0.215 & 0.303 & 0.424 \\
        RTI  & 0.017 & 0.178 & 0.080 & 0.013 & 0.233 & 0.088 & 0.063 & 0.167 & 0.222 \\
        Bronchitis & 0.013 & 0.177 & 0.096 & 0.023 & 0.256  & 0.101 & 0.122 & 0.183 & 0.259 \\
        Asthma & 0.006 & 0.130 & 0.061 & 0.014 & 0.349 & 0.167 & 0.170 & 0.202 & 0.296 \\
        \bottomrule[1.5pt]
    \end{tabular}
    }
    \caption{Correlation between diseases and the structured data extracted from clinical notes.}
    \label{tab:struct}
\end{table}

To find out the relation between diseases and test result fields and further show how the test results contribute to the diagnosis, the correlation between every pair of disease and test field is shown in Table~\ref{tab:struct}.
In the experiments, 19 test result fields are adopted and all of them come from blood routine and blood gas.
In Table~\ref{tab:struct}, ``Temp.'' represents the patient's temperature upon admission, HR, BR, SBP, and DBP represent the heart rate, breath rate, systolic blood pressure, and diastolic blood pressure of the patient.
SPO2, WBC, N, CRP, HGB, PLT, LY, BPH, PCO2, PO2, K, and Na represent the peripheral capillary oxygen saturation, white blood cells count, neutrophil percentage, C-reactive protein count, hemoglobin, platelet count, lymphocyte percentage, blood pH, partial pressure of carbon dioxide, partial pressure of oxygen, blood potassium, and blood sodium, respectively. 
According to Table~\ref{tab:struct}, a few blood test results, including HGB, BPH, K, and Na, have relatively significant correlations with the diseases, illustrating the reliability of using structured data as an assistance for disease identification.

\section{Discussion}
% no subheading
% use as conclusion

In this work, we have explored the possibility of using deep learning to effectively analyze raw clinical notes and identify respiratory diseases with similar symptoms in pediatrics, including pneumonia, RTI, bronchitis, and asthma.
A two-stage system is proposed to tackle the problem. It significantly outperforms traditional machine-learning-based methods and existing state-of-the-art deep-learning-based methods commonly used for clinical note analysis.
The system consists of two stages, test result structuralization and disease identification.
Visualization of gradients verifies that the proposed system is able to locate crucial information and learn inner-phrase relationships, increasing the interpretability of disease identification results. 

We have discovered that input data modality, CNN architecture and techniques for extracting fine-grained features are three major factors that contribute to the performance. 
(1) When using multi-modal data, the improvement in performance demonstrates the importance of maintaining the magnitude information of numerical data. 
(2) Conventional plain CNN architectures are unable to provide satisfactory results as deep plain CNNs cannot be effectively trained, while shallow CNNs are not powerful enough to fit the data. 
The infusion block uses the identity mapping to overcome the training problem and is able to significantly improve the performance by increasing the depth of the network. 
(3) Techniques including adaptive feature infusion and multi-modal attentive fusion use cross-modal information to attentively extract features. 
Max pooling in text data only keeps the most significant feature from each channel, and amplifies the differences between fine-grained features. 

In most EHR related tasks, the input consists of dense and abundant structuralized data, and the quality of such structuralized data has a major impact on the quality of results. 
The expert system designed by Badnjevic~\cite{badnjevic2018expert} needs to have all data available, and the system cannot function properly when necessary data are missing. 
Deepr system~\cite{wickramasinghe2017deepr} also uses clean and structured data as input, and the embedding system in Deepr takes EMRs in the ICD-10 format as input.
However, converting raw clinical notes into ICD-10 codes requires extensive effort.
In Deep EHR~\cite{liu2018deep}, the authors propose to use both full clinical notes and a large number of well-structuralized data fields, including numerical lab results, vital signs, and demographic data.
However, full clinical notes and a large number of structuralized data fields significantly increase the burden of both hospitals and patients by requiring them conduct hypothetico-deductive reasoning and many lab tests.
The diagnosis system proposed by Liang~\cite{liang2019evaluation} uses more structuralized fields and takes the reports from PACS (picture archiving and communication systems) as input, imposing the requirement of a specific medical device.
This method has been implemented and tested on our dataset, and its low mAP of 0.496 confirms that it cannot handle sparse structuralized data well, resulting in significant performance degradation.
In contrast, the proposed system in this paper only takes the chief complaint, physical examination results and clinical test results as input without requiring any modifications to the normal diagnosis procedure.

In the last few years, many works have adopted various deep learning algorithms~\cite{shickel2017deep, kwak2019deephealth}.
Wrenn~\cite{wrenn2005estimating}, who proposed to use a neural network to predict the length of stay of patients, became one of the pioneers in applying deep learning to EHR. 
Deep EHR~\cite{liu2018deep} explores the possibility of using CNNs and LSTMs to predict chronic diseases while Suresh~\cite{suresh2017clinical} tested the performance of CNNs and LSTMs in predicting a variety of clinical interventions. 
Chalapathy~\cite{chalapathy2016bidirectional} proposed to use a BLSTM with a conditional random field layer to extract clinical concepts from notes.
Wunnava~\cite{wunnava2018bidirectional} proposed to use a BLSTM for adverse drug event tagging.
These methods exploit the power of deep learning and achieve certain performance improvements over traditional machine learning based methods. 
Zhang~\cite{zhang2019modelling} also proposed to use a LSTM-based network to predict mortality, acute kidney injury and disease in the ICD-9 category. 

Recently, the attention mechanism in deep learning has also been introduced to EHR analysis to focus on automatically selected patterns and features. 
In ATTAIN~\cite{zhang2019attain}, an LSTM network with global, local, and flexible attention modules is used to predict disease progress. 
Pandey~\cite{pandey2017improving} proposed to use a recurrent neural network (RNN) with \textit{word2vec} and \textit{GloVe} embedding and attention to predict drug reactions.
Chu~\cite{chu2018using} proposed a neural-attention-based model to integrate the contextual information of words for adverse medical events detection. 
In BEHRT~\cite{li2020behrt}, the author proposed to use a transformer-based model similar to BERT~\cite{devlin2018bert} for multitask prediction and disease trajectory mapping. 
However, these works only incorporate the attention mechanism, but fail to exploit the distinctions between numerical and text data.
In the proposed system, we unleash the potential of the attention mechanism by adopting multi-modal input and proposing adaptive feature infusion and multi-modal attentive fusion to attentively merge multi-modal information and significantly improve the performance.

On the basis of the above findings and discussions, the performance of the proposed system may be further improved. 
The dataset we use consists of an estimated 14,000 clinical notes, and further increasing the dataset volume may further improve the performance of our system. 
Our structuralized data are highly sparse, and certain fields have over 90\% empty entries. 
Extracting numerical values more accurately from clinical notes and generating denser structured data should improve the performance. 
Techniques like pointwise group convolution~\cite{zhang2018shufflenet}, channel shuffling~\cite{zhang2018shufflenet}, and depthwise separable convolution~\cite{howard2017mobilenets} may be adopted to improve the efficiency of computation and memory consumption.

\section{Conclusions}

In this paper, we have explored the possibility of effectively analyzing unstructured clinical data and proposed a novel fine-grained system for diagnosing respiratory diseases in pediatrics. 
Unlike most of the previous researches in this field, the proposed system directly uses unstructured clinical data as input, automatically formats them in the result structuralization stage, and attentively exploits the multi-modal information from numerical lab results and text-form clinical notes in the disease identification stage. 
Thus, the system is easy for hospitals to adapt directly without changing the diagnostic process, reducing the burden of manually cleaning up and formatting the notes. 
The multi-modal information and attention-based modules in the system boost the disease identification performance of the system. 
Extensive experiments on a dataset with 14,697 clinical notes have proved the superiority of the proposed system in comparison to multiple competitive baseline methods. 

In the future, we will further improve the system in two potential directions.
First, we would like to integrate multiple tasks and more diseases into the system, enhancing the versatility. 
Second, it would be interesting to develop a more sophisticated method for numerical data extraction, improving the accuracy and compatibility of the result structuralization stage.

\section*{Data Availability}
The data that support the findings of this study were made available by the Children's Hospital of Zhejiang University School of Medicine under a license for the current study, and so are not publicly available. Data can however be available from the authors upon request with permission from the Children's Hospital of Zhejiang University School of Medicine. 

%\noindent LaTeX formats citations and references automatically using the bibliography records in your .bib file, which you can edit via the project menu. Use the cite command for an inline citation, e.g.  \cite{Hao:gidmaps:2014}.

%For data citations of datasets uploaded to e.g. \emph{figshare}, please use the \verb|howpublished| option in the bib entry to specify the platform and the link, as in the \verb|Hao:gidmaps:2014| example in the sample bibliography file.

\section*{Author contributions statement}

Gang Yu, Zheming Li, Yonggen Zhao, and Fenglei Sun collected, cleaned, formatted and analysed the data.
Zhongzhi Yu and Yemin Shi conceived and conducted the experiments and wrote the manuscript.
Xiaoqing Liu and Yizhou Yu analysed the results and revised the manuscript.
Yingshuo Wang, Yizhou Yu and Qiang Shu analysed the data and supervised the study.
All authors have reviewed and approved the manuscript.

\section*{Acknowledgements}

%Acknowledgements should be brief, and should not include thanks to anonymous referees and editors, or effusive comments. Grant or contribution numbers may be acknowledged.

\noindent Funding: This work was supported by the National Natural Science Foundation of China (No.62076218, No.81971616, No.62006004), the Zhejiang Province Research Project of Public Welfare Technology Application (LGF18H260004).

\section*{Declaration of Competing Interest}

The authors declare that they have no known competing financial interests or personal relationships that could have appeared to influence the work reported in this paper.

\section*{Appendix A. All Structuralized Fields}

\begin{table}[hp!]
    \centering
    \renewcommand\thetable{A.1}
    \resizebox{\columnwidth}{!}{
    \begin{tabular}{c|cccccccccc}
        \toprule[1.5pt]
        Item & \textbf{Gender} & \textbf{Age} & \textbf{Temp.} & \textbf{HR} & \textbf{BR} & \textbf{SBP} & \textbf{DBP} & Weight & \textbf{SPO2} & \textbf{WBC} \\
        \hline
        Density & \textbf{35.72\%} & \textbf{99.89\%} & \textbf{81.30\%} & \textbf{78.66\%} & \textbf{78.10\%} & \textbf{61.75\%} & \textbf{61.75\%} & 3.41\% & \textbf{7.59\%} & \textbf{72.97\%} \\
        \midrule[1.5pt]
        Item & GRA & \textbf{N} & \textbf{CRP} & MCHC & \textbf{HGB} & MCH & $\text{HGB}_A$ & MCV & RBC & PCV \\
        \hline
        Density & 3.17\% & \textbf{63.32\%} & \textbf{62.20\%} & 0.26\% & \textbf{68.15\%} & 0.13\% & 0.03\% & 0.35\% & 4.01\% & 0.53\% \\
        \midrule[1.5pt]
        Item & Retic & \textbf{PLT} & PCT & \textbf{LY} & $\text{LY}_A$ & AMS & Fbg & EOS & $\text{EOS}_A$ & $\text{PCT}_1$ \\
        \hline
        Density & 0.68\% & \textbf{68.23\%} & 0.02\% & \textbf{49.06\%} & 0.21\% & 0.37\% & 0.05\% & 0.49\% & 0.12\% & 2.67\% \\
        \midrule[1.5pt]
        Item & BASO & $\text{MONO}_A$ & MONO & ALT & AST & ALB & TP & TG  & BUREA & PUREA  \\
        \hline
        Density & 0.04\% & 0.02\% & 0.24\% & 3.57\% & 2.42\% & 0.87\% & 0.41\% & 0.28\% & 0.36\% & 0.33\%\\
        \midrule[1.5pt]
        Item& CR & \textbf{BPH} & PPH & \textbf{PCO2} & \textbf{PO2} & SO2 & FERR & TBIL  & \textbf{K} & \textbf{Na} \\
        \hline
        Density & 1.02\% & \textbf{10.59\%} & 0.05\% & \textbf{9.50\%} & \textbf{9.25\%} & 3.44\% & 0.03\% & 0.66\% & \textbf{9.72\%} & \textbf{8.57\%}  \\
        \midrule[1.5pt]
        Item& Lac & $\text{Ca}_2$ & TC & UAIC & BUN & C3C & C4 & BLO & PRO  & $\text{PC}_u$ \\
        \hline
        Density & 2.42\% & 3.67\% & 0.37\% & 0.06\% & 0.07\% & 0.19\% & 0.10\% & 4.43\% & 5.40\%  & 0.72\%\\
        \midrule[1.5pt]
        Item& $\text{PC}_s$ & $\text{PRBC}_{uL}$ & $\text{PRBC}_{Hp}$ & $\text{PRBC}_{No}$ & $\text{PRBC}_s$ & $\text{PWBC}_{uL}$ & $\text{PWBC}_{Hp}$  & $\text{PWBC}_{No}$ & $\text{PWBC}_{s}$& OB\\
        \hline
        % PC for PC1 
        Density & 0.20\% & 2.41\% & 1.52\% & 0.22\% & 0.28\% & 0.58\% & 1.03\% & 0.67\%  & 0.03\% & 0.61\%\\ 
        \midrule[1.5pt]
        Item & H24UPQ & MTP & MALB\\
        \hline
        Density & 0.24\% & 0.38\% & 0.02\% \\ 
        \bottomrule[1.5pt]
        
    \end{tabular}
    }
    \caption{All fields and their densities extracted in the test result structuralization stage. Selected fields passed into the disease identification stage are depicted in bold font.}
    \label{tab:all_struct}
\end{table}

\section*{Appendix B. Comparison of AP Results}

\begin{figure}[!htb]
    \centering
    \renewcommand\thefigure{B.1}
    \includegraphics[width=1.0\textwidth]{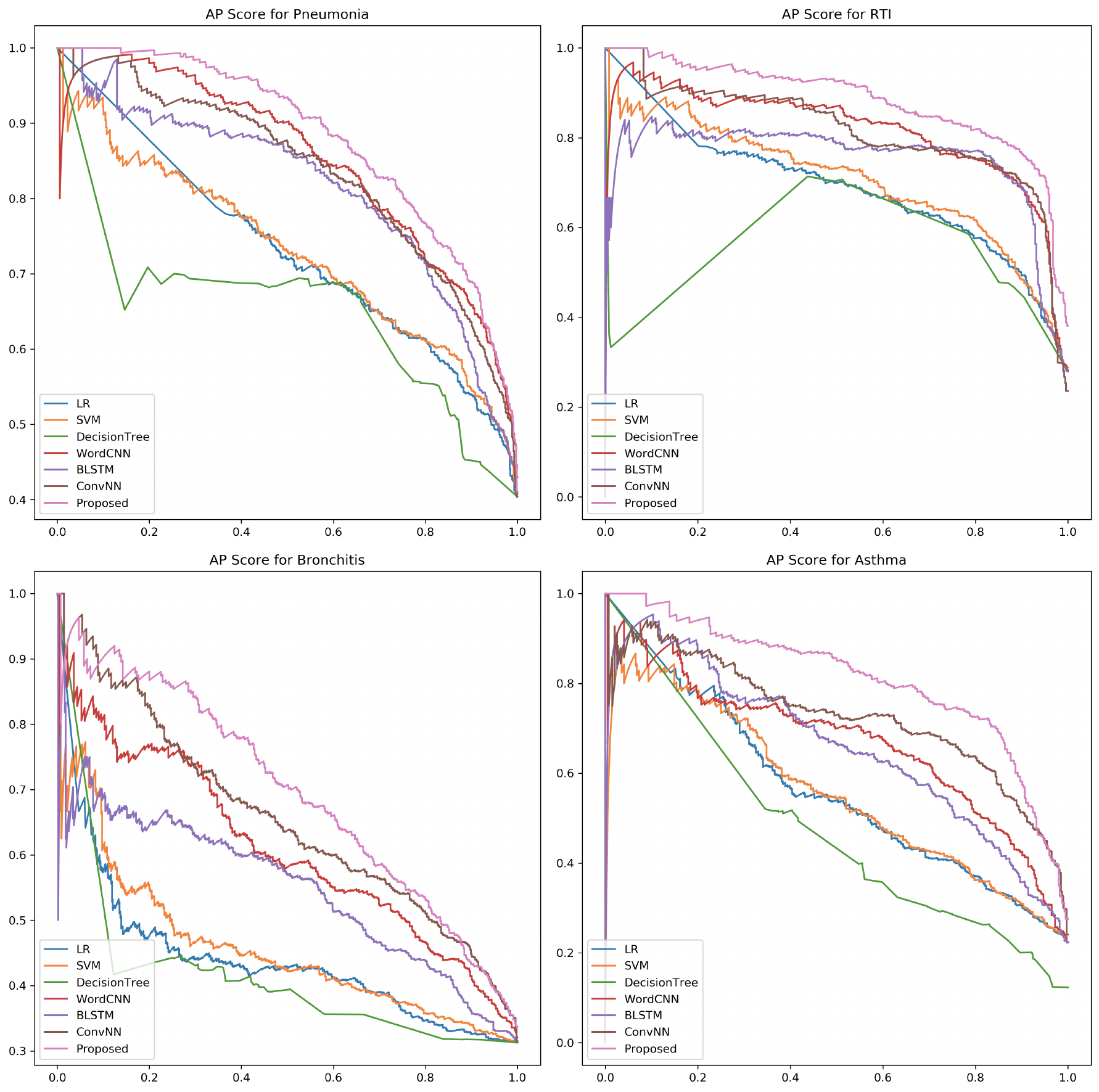}
    \caption{AP plots of the proposed and comparison methods on each of four respiratory diseases. }
    \label{fig:my_label}
\end{figure}

\bibliography{sample}

\end{document}